\documentclass[runningheads]{llncs}

\usepackage{eccv}

\usepackage{eccvabbrv}

\usepackage{graphicx}
\usepackage{booktabs}

\usepackage{mathtools}
\usepackage{adjustbox}
\usepackage{multirow}
\usepackage{colortbl}
\usepackage{makecell}
\usepackage{xcolor}

\usepackage[accsupp]{axessibility}  % Improves PDF readability for those with disabilities.

\usepackage[breaklinks,colorlinks,citecolor=eccvblue]{hyperref}
\usepackage{orcidlink}

\begin{document}

\title{VisReflect: Latent Visual Reflection for Fine-Grained Perception in Long Visual Context} 

\titlerunning{VisReflect}

\author{Xiaoqian Shen\orcidlink{0000-0001-6284-520X}\qquad
Mohamed Elhoseiny\orcidlink{0000-0001-9659-1551}}

\authorrunning{X. Shen et al.}

\institute{King Abdullah University of Science and Technology, Saudi Arabia\\
\email{\{xiaoqian.shen,mohamed.elhoseiny\}@kaust.edu.sa}\\
\url{https://xiaoqian-shen.github.io/VisReflect}}

\maketitle

\begin{abstract}
  Large Vision Language Models (LVLMs) have achieved remarkable success on vision–language tasks, yet fine-grained perception over high-resolution images and long-context videos remains challenging. As the number of visual tokens increases, the visual attention sink phenomenon becomes increasingly severe, causing irrelevant tokens to absorb a disproportionate amount of attention mass. Recent approaches attempt to mitigate this issue by explicitly predicting bounding boxes or temporal spans and re-encoding the cropped visual regions. Such methods depend on unreliable numeric localization in the discrete token space and incur significant computational overhead due to additional forward passes. In this work, we propose \textbf{VisReflect}, a simple yet effective framework that improves fine-grained perception in long visual contexts through latent visual reflection. Instead of decoding intermediate predictions into discrete tokens, the model generates continuous visual reflection that represents question-relevant visual features in the latent space. These reflections selectively emphasize salient regions or frames, guiding attention towards relevant visual tokens within a single forward pass. We conduct comprehensive evaluations on challenging high-resolution image benchmarks, including BLINK, V$^{*}$, and HRBench-4K/8K, as well as video understanding benchmarks such as MVBench, VideoMME, and MLVU. Our method consistently improves over strong baselines, achieving gains of $4.1\%$ on image benchmarks and $1.8\%$ on video benchmarks. Compared with zooming-based methods, our model achieves comparable performance while reducing inference time by roughly 44\% on video understanding.
\end{abstract}

\section{Introduction}
\label{sec:intro}

Large Vision Language Models (LVLMs) have recently achieved remarkable progress on advanced vision–language tasks, demonstrating strong performance in visual question answering and video reasoning. By integrating powerful language models with visual encoders, the models are able to reason over multimodal inputs and exhibit impressive capabilities in complex perception and reasoning scenarios. However, despite these successes, a persistent challenge remains: fine-grained perception in long visual contexts.

\begin{figure}[t]
    \centering
    \includegraphics[width=\textwidth]{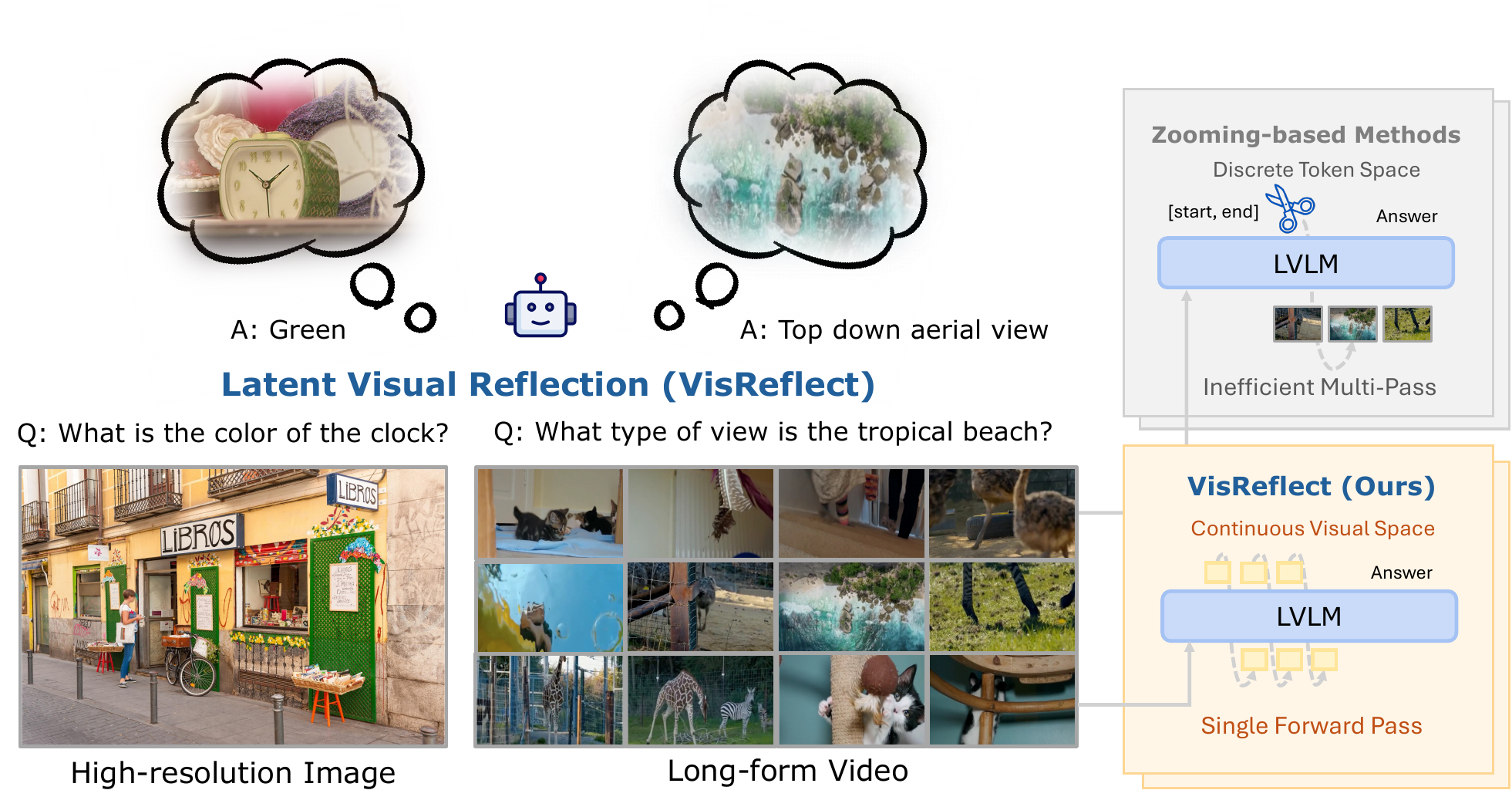}
    \caption{Existing zooming-based methods localize relevant regions or frames by predicting coordinates in discrete token space and performing repeated forward passes. In contrast, VisReflect generates latent visual reflection tokens in continuous visual space, enabling the model to internally recall question-relevant visual features within a single forward pass.}
    \label{fig:teaser}
\end{figure}

When processing high-resolution images, LVLMs must handle thousands of visual tokens to achieve fine-grained perceptual understanding. This challenge is further amplified in long-context video reasoning, where visual token sequences grow substantially. Ideally, the model should selectively attend to visual information that is most relevant to the textual query. However, as the number of visual tokens increases, the \emph{Visual Attention Sink}~\cite{kang2025see} phenomenon becomes more severe, causing irrelevant tokens to absorb a disproportionate amount of attention mass. Consequently, the model struggles with small-object perception in high-resolution images or temporally localized reasoning in long-form videos.

To address this challenge, recent studies have introduced the ``\emph{Thinking-with-Images/Videos}'' paradigm~\cite{zhang2025thyme,zheng2025deepeyes,pan2025timesearch,fu2025love,shen2025zoom}, which aims to preserve global context while capturing fine-grained details in long visual sequences. Specifically, these methods prompt the model to predict the bounding box of a region of interest or the temporal span of salient frames. Based on these predictions, a second fine-grained forward pass is performed, where visual tools are invoked to crop or zoom into the selected image regions or video segments, and the refined inputs are re-encoded for detailed understanding. Although this paradigm yields performance improvements, it exposes two critical limitations. \textbf{(i) Reliability of numeric predictions.} The model generates deterministic tokens to represent bounding box coordinates or temporal intervals, which are expressed purely in the textual space and do not faithfully model the underlying visual feature space. As a result, the second forward pass is highly sensitive to localization errors. \textbf{(ii) Increased computational cost due to additional prefilling.} Iterative cropping or zooming requires the visual features to be re-forwarded to the language model, substantially increasing inference latency and computational overhead.

These limitations lead us to rethink, how should LVLMs retrieve relevant visual evidence from massive token sequences. Instead of explicitly zooming into pixel-space regions through discrete token predictions, \emph{can a model directly recall relevant visual representations within its latent space?}

Motivated by this perspective, we propose \textbf{VisReflect}, a simple yet effective framework that enhances fine-grained perception in high-resolution images and long-context videos without explicit coordinate-based cropping or temporal zooming. Instead of predicting discrete localization tokens, it generates continuous visual reflection embeddings that recall the visual features of salient regions or frames in the latent space. These reflection embeddings guide attention toward informative visual tokens, enabling the model to focus on relevant regions or frames within a single forward pass. As a result, our approach provides a computationally efficient alternative to multi-pass zooming strategies, offering a unified solution for long-context visual understanding.

We evaluate our method on challenging high-resolution image benchmarks, including BLINK~\cite{fu2024blink}, V$^{*}$, and HRBench~\cite{wang2025divide}, as well as on long-video benchmarks such as VideoMME~\cite{fu2025video} and MLVU~\cite{zhou2025mlvu}, which demand precise small-object perception and temporally localized segment reasoning. Experimental results demonstrate an average improvement of $4.1\%$ over the base model on image benchmarks and a $1.8\%$ gain over strong baselines on video benchmarks. Moreover, our VisReflect achieves comparable performance than multi-pass ``thinking-with-video'' approaches while reducing inference time by approximately 44\%.

\noindent\textbf{Contribution.} Our contribution is summarized as follows:

\begin{itemize}
    \item We propose \textbf{VisReflect}, a novel framework that reflects question-relevant visual information from long visual contexts directly in the continuous embedding space, avoiding unreliable numeric localization and costly multi-pass inference required by prior tool-using methods.

    \item We analyze the attention maps from the visual reflection tokens to the original visual inputs. The reflected visual representations guide the model to focus its attention on informative regions or frames within large visual contexts, mitigating the attention sink caused by massive visual tokens.
    
    \item Comprehensive experiments demonstrate that VisReflect improves fine-grained perception in both high-resolution images and long-form videos, achieving gains of \textbf{+4.1\%} and \textbf{+1.8\%}. It achieves comparable performance to zooming-based methods that require multiple forward passes, while reducing video inference time by \textbf{44\%}.
\end{itemize}

\section{Related Work}
\label{sec:relate}

\subsection{Thinking with Images} 

Large vision language models (LVLMs)~\cite{liu2023visual,zhu2023minigpt} have recently achieved strong performance across a wide range of visual understanding and reasoning tasks. Building upon these foundations, several studies~\cite{chen2023minigptv2largelanguagemodel,peng2023kosmos,chen2023shikra,rasheed2024glamm} enhance spatial perception by enabling models to predict explicit coordinate outputs, allowing them to localize small or fine-grained objects within images.
More recently, research on fine-grained multimodal reasoning has moved toward a “Thinking-with-Images” paradigm, in which models dynamically retrieve localized visual evidence during inference rather than depending exclusively on a single global image representation~\cite{yu2025zoom, wei2025perception,zheng2025deepeyes,lai2025mini,zhang2025thyme,wang2025pixel}.
In this context, DeepEyes~\cite{zheng2025deepeyes} and Mini-o3~\cite{lai2025mini} employ reinforcement learning to incentivize the usage of visual tools such as zooming-in, cropping or searching, while Thyme~\cite{zhang2025thyme} and PixelReasoner~\cite{wang2025pixel} enable models to generate code or perform pixel-space operations to manipulate visual inputs dynamically.
Although effective, this paradigm incurs substantial inference overhead due to repeated visual encoding passes or iterative tool-calling loops, leading to increased latency. Moreover, these methods are constrained to predict numeric bounding box coordinates in the textual space, which do not truly reflect the underlying visual features. In our work, we encourage the model to perform latent visual reflection by generating visual representations in the LLM’s output space that align with past salient region features. This helps the model re-attend to relevant information within the massive visual token sequence of high-resolution images, without explicit cropping during inference that would otherwise increase computational cost.

\subsection{Thinking with Videos}
Large vision-language models process videos by extracting and encoding frames, and then concatenating them into final video representations. To address massive number of visual tokens for long videos, several works train on sparsely sampled frames~\cite{li2023videochat,ataallah2024minigpt4,cheng2024videollama}, while others try to handle long videos by token pooling~\cite{maaz2023videochatgpt,li2023llama,song2023moviechat}, token compression~\cite{shen2024longvu} or segment retrieval~\cite{ataallah2024goldfish,shen2025vgent}. More recently, a new paradigm ``\textit{Thinking-with-Video}'' has emerged~\cite{fu2025love, shen2025zoom, yan2025videochat, pan2025timesearch}. The model first generates intermediate temporal cues to identify query-relevant segments, and then performs a second forward pass by “zooming in” on these salient frames. This test-time scaling strategy effectively preserves global context while refining local details through re-inference. Despite its effectiveness, the reliance on multiple forward passes substantially increases inference latency and computational cost. In contrast, we propose a latent visual reflection mechanism that achieves comparable performance improvements within a single forward pass.

\subsection{Latent Reasoning}

Latent reasoning~\cite{coconut,shen2025codi,cheng2024compressed,li2025latent} performs intermediate reasoning entirely in the latent space, without explicitly decoding intermediate steps into natural language. Instead of generating and appending reasoning tokens, the model performs reasoning by recursively using the latent state from the previous iteration as the input embedding to the next, and returns to text space only at the final step to generate the answer. 
This paradigm, instead of converting continuous embeddings into a deterministic token space, enabling reasoning beyond language and exploiting rich, high-dimensional representations. In this work, we leverage the continuous latent embeddings to align with salient visual features, enhancing the model's perceptual capabilities in long visual contexts such as high-resolution images or long video sequences.

\section{Method}
\label{sec:method}

\subsection{Preliminaries}

A Large Vision-Language Model (LVLM) consists of three components: 
(i) a vision encoder $\mathcal{V}$, 
(ii) a projection module $\mathcal{P}$ that maps visual representations into the language embedding space, and 
(iii) a large language model $\mathcal{M}$.

Let $\mathbf{I}$ denote an input image and $\mathbf{V}=\{\mathbf{I}_t\}_{t=1}^{T}$ denote a video with $T$ frames.
Given a visual input (image or video) together with a textual query $\mathbf{Q} = \{q_i\}_{i=1}^{L_q}$ of length $L_q$, the vision encoder extracts visual tokens, which are projected into the language embedding space as:
\begin{equation}
\mathbf{Z}_v = \mathcal{P}(\mathcal{V}(\mathbf{I})) 
\quad \text{or} \quad
\mathbf{Z}_v = \mathcal{P}(\mathcal{V}(\mathbf{V})).
\end{equation}
Here, $\mathbf{Z}_v \in \mathbb{R}^{N \times d}$ denotes $N$ visual token embeddings of dimension $d$. 
For video inputs, $N = T \times N_s$, where $N_s$ is the number of spatial tokens per frame.

The textual query is embedded via a token embedding function $\mathcal{E}(\cdot)$:
\begin{equation}
\mathbf{Z}_q = \{\mathcal{E}(q_i)\}_{i=1}^{L_q}, 
\qquad \mathbf{Z}_q \in \mathbb{R}^{L_q \times d}.
\end{equation}

The concatenated sequence $[\mathbf{Z}_v; \mathbf{Z}_q]$ is fed into the language model $\mathcal{M}$, which autoregressively generates an answer sequence $\mathbf{Y}=\{y_i\}_{i=1}^{L_y}$:
\begin{equation}
p(\mathbf{Y} \mid \mathbf{I}, \mathbf{Q})
= \prod_{i=1}^{L_y} 
p(y_i \mid y_{<i}, \mathbf{Z}_v, \mathbf{Z}_q).
\end{equation}

Let $\mathbf{h}_i \in \mathbb{R}^{d}$ denote the last-layer hidden state of $\mathcal{M}$ at decoding step $i$. 
The token distribution is computed through a language modeling head $\mathcal{W}_{\mathrm{LM}}$:
\begin{equation}
p(y_{i+1} \mid y_{\le i})
= \mathrm{Softmax}(\mathcal{W}_{\mathrm{LM}} \mathbf{h}_i).
\end{equation}

\paragraph{Limitation.}
When the visual input are high-resolution images or long videos, the number of visual tokens $N$ becomes large. 
The self-attention mechanism allocates attention across all tokens, reducing effective focus on query-relevant regions and impairing fine-grained perception.

Existing grounding approaches predict explicit coordinates, such as bounding boxes $[x_1,y_1,x_2,y_2]$ for images or temporal intervals $[t_s,t_e]$ for videos. 
However, these predictions are constrained to the discrete token space and therefore do not faithfully capture the underlying visual feature space.
Moreover, such methods typically require cropping and re-encoding the selected region in an additional forward pass, increasing computational overhead and breaking strict end-to-end differentiability.

\begin{figure}[t]
    \centering
    \includegraphics[width=\textwidth]{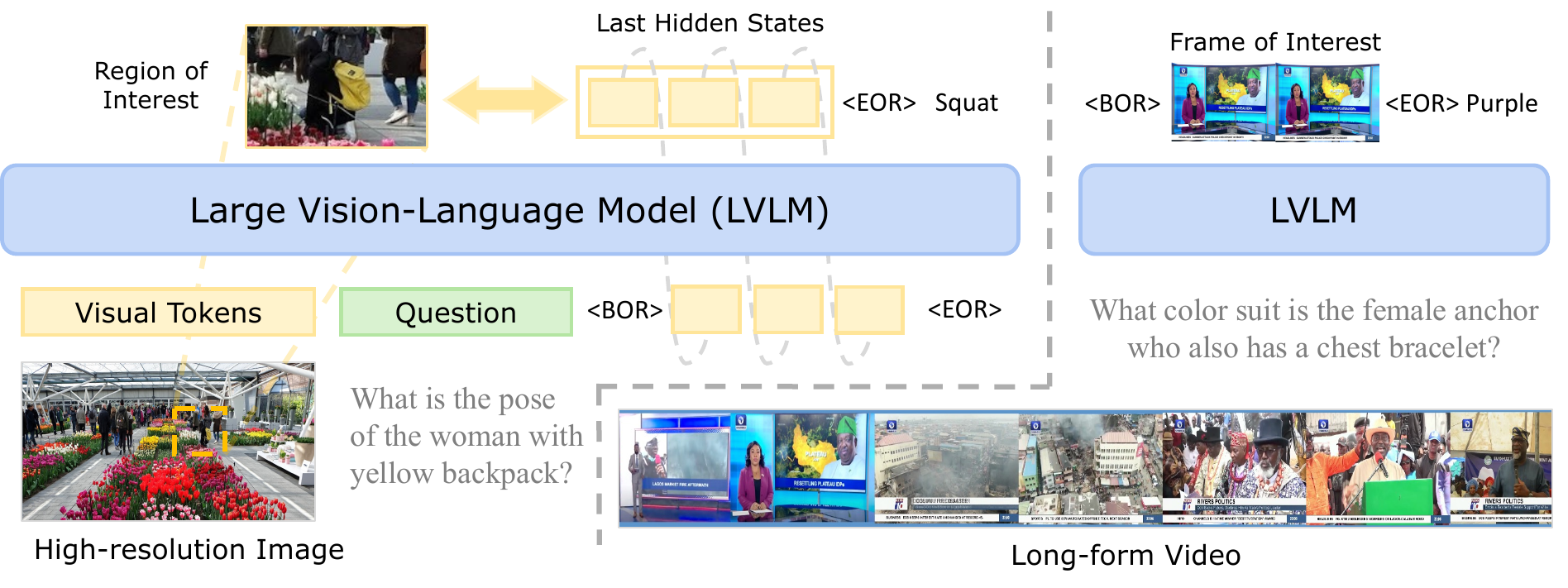}
    \caption{\textbf{Overview of VisReflect}. Given visual tokens extracted from a high-resolution image or a long-form video together with a textual query, a LVLM generates a sequence of visual reflection tokens between the special tokens \texttt{<BOR>} and \texttt{<EOR>}. These tokens are trained to approximate the latent visual representations of the region of interest (for images) or frames of interest (for videos) that are relevant to the question. Instead of explicitly cropping images or re-encoding selected frames, the model recalls question-relevant visual features directly in the latent embedding space using its last hidden states. The reflected visual features then guide the model’s attention toward informative parts of the visual context, enabling accurate fine-grained perception in high-resolution images and long-form videos while requiring only a single forward pass.}
    \label{fig:main}
\end{figure}

\subsection{VisReflect}

To overcome these limitations, we introduce a \emph{latent visual reflection} mechanism that enables the model to reflect question-relevant visual information from long visual contexts before producing the final answer. Concretely, the model generates a sequence of latent visual embeddings that reside in the same embedding space as the visual input, allowing them to directly represent salient regions or frames within the visual context. Unlike prior approaches that rely on predicting numeric coordinates for cropping or zooming, our method reflects the relevant visual content in the continuous embedding space within a single forward pass.

More specifically, we add three special tokens to the vocabulary:
$\texttt{<BOR>}$ (beginning of reflection), 
$\texttt{<VR>}$ (visual reflection token), and 
$\texttt{<EOR>}$ (end of reflection).
These tokens are designed to trigger the visual reflection mode within LVLMs, guiding the model to internally generate embeddings that reflect the relevant visual features, effectively recalling past visual content within the latent space.

\subsubsection{Visual Reflection for Image Understanding.}

Let $\mathbf{Z}_v = \{\mathbf{z}_n\}_{n=1}^{N}$ denote the spatial visual tokens extracted from an input image, where $\mathbf{z}_n \in \mathbb{R}^{d}$.

Given a bounding box annotation, we identify the indices of the $K$ visual tokens that fall inside the region of interest (ROI), denoted as ${i_1, i_2, \dots, i_K}$, where $K$ is the number of tokens within the ROI.

During training, the model generates a reflection sequence:
$$
\texttt{<BOR>} \ 
\texttt{<VR>}_1 \ 
\texttt{<VR>}_2 \ 
\dots \ 
\texttt{<VR>}_K \ 
\texttt{<EOR>}.
$$

Let $\mathbf{h}^{\mathrm{VR}}_k \in \mathbb{R}^{d}$ denote the last-layer hidden state corresponding to the $k$-th $\texttt{<VR>}$ token. 
We enforce a one-to-one alignment between reflection embeddings and ROI visual tokens by maximizing their cosine similarity:
\begin{equation}
\mathcal{L}_{\mathrm{VR}}^{\mathrm{img}}
=
\frac{1}{K} \sum_{k=1}^{K} 
\Bigg(
1 - 
\frac{\mathbf{h}^{\mathrm{VR}}_k \cdot \mathbf{z}_{i_k}}
     {\|\mathbf{h}^{\mathrm{VR}}_k\|_2 \, \|\mathbf{z}_{i_k}\|_2}
\Bigg)
\end{equation}

Intuitively, the generated visual reflection embeddings act as latent anchors. Since these embeddings approximate visual tokens corresponding to salient regions, they naturally propagate attention mass from the reflection tokens toward the associated visual tokens. As a result, the model implicitly re-focuses its attention distribution on question-relevant subsets of the visual context, alleviating the attention sink caused by large numbers of visual tokens.

\subsubsection{Visual Reflection for Video Understanding.}

For a video input, the encoder produces frame-wise spatial tokens:
\begin{equation}
\mathbf{Z}_v = \{\mathbf{Z}^{(t)}\}_{t=1}^{T},
\qquad
\mathbf{Z}^{(t)} = \{\mathbf{z}^{(t)}_n\}_{n=1}^{N_s},
\end{equation}
where $N_s$ denotes the number of spatial tokens per frame. Given temporal boundaries $[t_s, t_e]$, we uniformly sample $J$ frames with indices $t_1, t_2, \dots, t_J$, where each $t_j \in [t_s, t_e]$.

For each sampled frame, we recall all spatial tokens. 
Therefore, the model generates $J\times N_s$ tokens for visual reflection.
$$
\texttt{<BOR>} \ 
\texttt{<VR>}_1 \ 
\texttt{<VR>}_2 \ 
\dots \ 
\texttt{<VR>}_{J\times N_s} \ 
\texttt{<EOR>}.
$$

The salient frame feature alignment is defined as:
\begin{equation}
\mathcal{L}_{\mathrm{VR}}^{\mathrm{vid}}
=
\frac{1}{J \times N_s}
\sum_{j=1}^{J}
\sum_{n=1}^{N_s}
\Bigg(
1 - 
\frac{
\mathbf{h}^{\mathrm{VR}}_{(j-1)\times N_s+n} \cdot \mathbf{z}^{(t_j)}_{n}
}{
\|\mathbf{h}^{\mathrm{VR}}_{(j-1)\times N_s+n}\|_2 \, 
\|\mathbf{z}^{(t_j)}_{n}\|_2
}
\Bigg).
\end{equation}

\subsubsection{Language Mode v.s. Visual Reflection Mode.} The model operates in two modes. In text generation mode, the last hidden states are projected to vocabulary logits through the language model's head, the embedding of the next input token is then set to the embedding of the generated token:
\begin{equation}
p(y_{i+1} \mid y_{\le i})
= \mathrm{Softmax}(\mathcal{W}_{\mathrm{LM}} \mathbf{h}_i),
\quad
\mathbf{E}_{i+1} = \mathcal{E}(y_{i+1}).
\end{equation}

In visual reflection mode, we use the last hidden state from the previous token to replace the input embedding:
\begin{equation}
\mathbf{E}_{i+1} = \mathbf{h}^{\mathrm{VR}}_i, \quad \mathrm{if} \, y_i = \texttt{<VR>}
\end{equation}

The final answer is then generated conditioned on the continuous visual reflection, which emphasizes the regions or frames most relevant to the query, implicitly recalling fine-grained information from the long visual context.

\subsubsection{Training Objective.} The overall objective combines language modeling and visual alignment, $\mathcal{L} = 
\mathcal{L}_{\mathrm{LM}} + \lambda \mathcal{L}_{\mathrm{VR}}$, where $\mathcal{L}_{\mathrm{LM}}$ is the standard cross-entropy loss for answer generation, 
$\mathcal{L}_{\mathrm{VR}} \in \{\mathcal{L}_{\mathrm{VR}}^{\mathrm{img}}, \mathcal{L}_{\mathrm{VR}}^{\mathrm{vid}}\}$ depends on modality, 
and $\lambda$ balances linguistic supervision and latent visual alignment.

The entire framework is trained end-to-end in a single forward pass, enabling computational efficiency and enhancing attention to relevant regions in high-resolution images and long-form videos. Localization annotations are used only as a training signal, and the model remains fully coordinate-free during inference.

\subsubsection{Inference.} During inference, we append \texttt{<BOR>} to the end of the question prompt. Then the model naturally generates \texttt{<VR>} tokens following the training format. The hidden state of \texttt{<VR>} token will replace its token embedding for the next decoding step. Once the number of generated \texttt{<VR>} tokens reaches the limit, we force the next token to be \texttt{<EOR>} to terminate the reflection process.

\section{Experiments}
\label{sec:exp}

\subsection{Experimental Setup}

\noindent\textbf{Benchmarks and Metrics.}
We conduct comprehensive evaluations on both high-resolution image benchmarks and long video understanding benchmarks.

\paragraph{Image Understanding.}
We evaluate our method on BLINK~\cite{fu2024blink}, HRBench-4K~\cite{wang2025divide}, HRBench-8K~\cite{wang2025divide}, and V$^{*}$~\cite{wu2024v}. BLINK~\cite{fu2024blink} contains perception-demanding tasks such as multi-view reasoning and relative depth estimation. V$^{*}$~\cite{wu2024v} and HRBench~\cite{wang2025divide} focus on fine-grained attribute recognition and spatial reasoning under high-resolution settings, where target objects typically occupy only a small fraction of the image. Such setups pose significant challenges due to the large number of visual tokens and the need for precise regional perception.

\paragraph{Video Understanding.}
For long-video understanding, we evaluate our method on MLVU~\cite{zhou2025mlvu}, MVBench~\cite{li2024mvbench}, and VideoMME~\cite{fu2025video}. These benchmarks require temporal reasoning over extended contexts, testing the model's ability to identify relevant actions or events within long video sequences. We report multi-choice accuracy as the evaluation metric.
\newline
\newline
\noindent\textbf{Baselines.} We adopt Qwen-2.5-VL-7B~\cite{bai2025qwen25vl} as the backbone and primary baseline, and further compare our approach against several recent state-of-the-art models for image and video understanding.

\paragraph{Image Understanding.} For image-level evaluation, we compare our method with recent visual reasoning approaches, including PixelReasoner~\cite{wang2025pixel}, PAPO~\cite{wang2025perception}, Thyme-VL~\cite{zhang2025thyme}, and DeepEyes~\cite{zheng2025deepeyes}. Notably, Thyme-VL~\cite{zhang2025thyme} and DeepEyes~\cite{zheng2025deepeyes} follow a ``thinking-with-images'' paradigm, which involves cropping the image and performing re-inference on selected regions, thereby requiring multiple forward passes. For fair comparison, we additionally report their performance without tool usage (i.e., single-pass inference), evaluating the models' standalone capabilities without additional inference.

\paragraph{Video Understanding.}
For video understanding, we compare against strong video-language models, including LLaVA-OneVision~\cite{li2024llavaonevision}, LongVU~\cite{shen2024longvu}, and VideoChat-R1~\cite{li2025videochat}. Furthermore, we evaluate against ``thinking-with-video'' approaches that first perform temporal grounding with respect to the query, then zoom into salient frames for a second-stage inference. This category includes LOVE-R1~\cite{fu2025love}, VideoChat-R1.5~\cite{yan2025videochat}, and TimeSearch-R~\cite{pan2025timesearch}. These methods similarly rely on multi-stage processing, in contrast to our single-pass visual reflection framework.
\newline
\newline
\noindent\textbf{Implementation Details.}
We adopt Qwen-2.5-VL-7B~\cite{bai2025qwen25vl} as the backbone model. Training contains an image stage followed by a video stage.
\paragraph{Stage 1: Image Training.}
We train on image data with the maximum input resolution set to $4096 \times 28 \times 28$ pixels and the minimum resolution set to $128 \times 28 \times 28$ pixels, corresponding to a visual token range of $128 \leq N \leq 4096$. We use the VisualCoT~\cite{shao2024visual} as image training data and filter the training samples to retain only those whose bounding box area is smaller than 10\% of the entire image.
\paragraph{Stage 2: Video Training.}
We initialize from the image-stage checkpoint and continue training on video data. The maximum number of visual tokens is set to 16,384, while the maximum pixel resolution per frame is constrained to $128 \times 28 \times 28$. We sample $J=3$ visual reflection frames from the annotated temporal span. The training datasets include QVHighlights~\cite{lei2021detecting}, ActivityNet~\cite{yu2019activitynet}, NExT-GQA~\cite{xiao2024can}, and the multiple-choice split of PLM-Video~\cite{cho2025perceptionlm}. We further restrict the temporal clue span to be shorter than 10 seconds to emphasize localized reasoning.

In both stages, the visual encoder and multimodal projector are kept frozen, and only the parameters of the language model are updated. The learning rate is set to $1\times10^{-5}$ with a warmup ratio of 0.03. We set $\lambda = 0.1$ to weight the alignment loss. All models are trained on 8 NVIDIA H100 GPUs (80GB).

\begin{table*}[t]
\centering
\begin{adjustbox}{width=\linewidth,center}
\renewcommand{\arraystretch}{1.05}
\begin{tabular}{lcccccc}
\toprule  \multirow{2}{*}{\textbf{Models}}  & \multirow{2}{*}{\textbf{BLINK}} & \multirow{2}{*}{\textbf{HRBench-4K}} & \multirow{2}{*}{\textbf{HRBench-8K}} & \multicolumn{3}{c}{\textbf{V$^{*}$}} \\
\cmidrule(lr){5-7} 
& & & & Attribute & Spatial & Overall \\
\midrule
\multicolumn{7}{c}{\textbf{\textit{Multiple Prefilling}}}\\
\midrule
\rowcolor{gray!15} Thyme-VL~\cite{zhang2025thyme} & - & 77.0 & 72.0 & 83.5 & 80.3 & 82.2 \\
\rowcolor{gray!15} DeepEyes~\cite{zheng2025deepeyes} & - & 75.1 & 72.6 & 91.3 & 88.2 & 90.1 \\
\midrule
\multicolumn{7}{c}{\textbf{\textit{Single Prefilling}}}\\
\midrule
Qwen2.5-VL~\cite{bai2025qwen25vl} & 56.5 & 70.9 & 66.9 & 80.8 & 76.3 & 79.1 \\
% Qwen2.5-VL Finetune & 55.2 & 71.3 & 65.9 & 80.1 \\
PixelReasoner~\cite{wang2025pixel} & 56.1 & 71.2 & 64.2 & 81.7 & 76.3 & 79.6 \\
PAPO~\cite{wang2025perception} & 55.6 & 70.5 & 69.1 & 80.8 & 77.6 & 79.5 \\
Thyme-VL$^{\dagger}$~\cite{zhang2025thyme} & 55.2 & 72.6 & 67.8 & 81.7 & 78.9 & 80.6 \\
DeepEyes$^{\dagger}$~\cite{zheng2025deepeyes} & 56.5 & 71.2 & 68.3 & 82.6 & 76.3 & 80.1 \\
LVR~\cite{li2025latent} & 55.3 & 71.4 & 67.2 & 81.7 & 79.0 & 80.6 \\
\rowcolor{eccvblue!15} VisReflect (Ours) & \textbf{57.1} & \textbf{73.8} & \textbf{70.1} & \textbf{86.7} & \textbf{82.3} & \textbf{84.9} \\
\bottomrule
\end{tabular}
\end{adjustbox}
\caption{\textbf{Performance on image understanding benchmarks.} $\dagger$ denotes a single prefilling without cropping. Their default tool-using results are marked in gray.}
\label{tab:image}
\end{table*}

\subsection{Visual Reflection for Fine-Grained Image Perception}

Table~\ref{tab:image} demonstrates the performance for fine-grained image perception.

\noindent\textbf{Comparison with the backbone model.}
Compared with the backbone model Qwen2.5-VL, our method achieves consistent improvements across all benchmarks, including BLINK (+0.6), HRBench-4K (+2.9), HRBench-8K (+3.2), and V$^{*}$ (+5.8). These gains indicate that visual reflection effectively enhances the model's ability to focus on question-relevant regions in high-resolution images.
\newline
\newline
\noindent\textbf{Comparison with state-of-the-art baselines.}
Our method outperforms the strongest runner-up (Thyme-VL$^{\dagger}$) on HRBench-4K (+1.2), HRBench-8K (+2.3), and on V$^{*}$ (+4.3), demonstrating the advantage of reflecting visual information in the latent space for fine-grained perception.
\newline
\newline
\noindent\textbf{Comparison with region-cropping methods.}
Our method approaches the performance of multi-pass cropping-based models such as Thyme-VL and DeepEyes. This highlights that latent visual reflection can serve as an effective alternative to explicit zoom-in operations for long visual context understanding while reducing computational overhead.

\begin{table*}[t]
\centering
\begin{adjustbox}{width=\linewidth,center}
\renewcommand{\arraystretch}{1.05}
\begin{tabular}{lccc}
\toprule  \textbf{Models} & \textbf{MLVU} & \textbf{MVBench} & \textbf{VideoMME (w/o \& w/ sub.)} \\
\midrule
LLaVA-OneVision~\cite{li2024llavaonevision} & 64.7 & 56.7 & 58.2 / 61.5 \\
LongVU~\cite{shen2024longvu} & 65.4 & 66.9 & 60.6 / 63.7 \\
LongVILA~\cite{chen2024longvila} & 67.0 & 67.1 & 60.1 / 65.1 \\
NVILA~\cite{liu2025nvila} & 70.1 & 68.1 & 64.2 / 70.0 \\
VideoChat-R1~\cite{li2025videochat} & 69.5 & 67.9 & 64.3 / 69.1 \\
Qwen2.5-VL~\cite{bai2025qwen25vl} & 69.8 & 68.4 & 65.2 / 70.7 \\
\rowcolor{eccvblue!15} VisReflect (Ours) & \textbf{70.6} & \textbf{70.4} & \textbf{67.4} /	\textbf{72.7} \\
\bottomrule
\end{tabular}
\end{adjustbox}
\caption{\textbf{Performance on video understanding benchmarks.}}
\label{tab:video}
\end{table*}

\begin{table*}[t]
\centering
\begin{adjustbox}{width=\linewidth,center}
\renewcommand{\arraystretch}{1.05}
\begin{tabular}{lcccccc}
\toprule  \multirow{2}{*}{\textbf{Models}} & \multicolumn{2}{c}{\textbf{MLVU}} & \multicolumn{2}{c}{\textbf{MVBench}} & \multicolumn{2}{c}{\textbf{VideoMME (w/o sub.)}} \\
\cmidrule(lr){2-3} \cmidrule(lr){4-5} \cmidrule(lr){6-7}
& Acc$\uparrow$ & Time$\downarrow$ & Acc$\uparrow$ & Time$\downarrow$ & \qquad Acc$\uparrow$ & \quad Time$\downarrow$ \\
\midrule
\multicolumn{7}{c}{\textbf{\textit{Multiple Prefilling }}}\\
\midrule
LOVE-R1~\cite{fu2025love} & 67.4 & 32.5s & 66.6 & 1.4s & \qquad 66.2 & 29.8s \\
VideoChat-R1.5-M~\cite{yan2025videochat} & 70.9 & 37.1s & 70.6 & 1.7s & \qquad 67.1 & 30.0s \\
TimeSearch-R~\cite{pan2025timesearch} & 71.5 & 32.8s & 69.8 & 1.6s & \qquad 66.6 & 28.5s\\
\midrule
\multicolumn{7}{c}{\small{\textbf{\textit{Single Prefilling}}}}\\
\midrule
\rowcolor{eccvblue!15} VisReflect (Ours) & 70.6 & 18.5s & 70.4 & 0.9s & \qquad 67.3 & 16.5s \\
\bottomrule
\end{tabular}
\end{adjustbox}
\caption{\textbf{Comparison with zooming-based video models.} Our method achieves comparable performance to zooming-based approaches that require multiple forward passes, while reducing inference time by 44\%.}
\label{tab:zoom}
\end{table*}

\subsection{Visual Reflection for Long Video Understanding}

Long videos pose a more severe challenge: as the number of frames increases, attention over all tokens becomes increasingly distracting, making it difficult for models to focus on the most salient moments.
\newline
\newline
\noindent\textbf{Comparison with state-of-the-art baselines.} As reported in Table~\ref{tab:video}, our approach consistently outperforms strong baselines on long video understanding benchmarks, achieving gains of +0.8 on MLVU~\cite{zhou2025mlvu}, +2.0 on MVBench~\cite{li2024mvbench}, and +2.1/+2.0 on VideoMME~\cite{fu2025video} w/o or with subtitles. These improvements demonstrate that visually reflecting on target frames enables the model to preserve global context while capturing critical detail within a single forward pass.

\noindent\textbf{Comparison with frame-zooming methods.} Notably, experimental results in Table~\ref{tab:zoom} show that our latent visual reflecting achieves performance comparable to ``Thinking-with-Video'' approaches that require multiple forward passes by zooming into salient frames, while substantially reducing inference latency.

\subsection{Ablation Studies}

\noindent\textbf{Effect of Visual Reflection.} Table~\ref{tab:ablation} presents the contribution of visual reflection. Fine-tuning Qwen2.5-VL alone yields only limited improvements, while incorporating visual reflection significantly boosts performance. We further compare mean squared error (MSE) and cosine similarity (COS) as the alignment loss, and observe that COS provides more stable training and consistently better performance.

\begin{table*}[h]
\centering
\begin{adjustbox}{width=\linewidth,center}
\renewcommand{\arraystretch}{1.05}
\begin{tabular}{lcccc}
\toprule  \textbf{Models}  & \textbf{BLINK} & \textbf{HRBench-4K} & \textbf{HRBench-8K} & \textbf{V$^{*}$} \\
\midrule
Qwen2.5-VL~\cite{bai2025qwen25vl} & 56.5 & 70.9 & 66.9 & 79.1 \\
Qwen2.5-VL Finetune & 55.2 & 71.3 & 65.9 & 80.1 \\
+ Visual Reflection (MSE) & 56.4 & 72.8 & 69.3 & 83.2 \\
\rowcolor{eccvblue!15} + Visual Reflection (COS) & 57.1 & 73.8 & 70.1 & 84.9\\
\bottomrule
\end{tabular}
\end{adjustbox}
\caption{\textbf{Effect of visual reflection and alignment loss choice.} We compare different loss functions for aligning the generated visual reflection tokens with target visual features.}
\label{tab:ablation}
\end{table*}

\noindent\textbf{Number of \texttt{<VR>} Tokens during Training.} We study how the number of generated \texttt{<VR>} tokens affects model performance. During training, we control the maximum number of visual reflection tokens that the model is required to generate. As shown in Table~\ref{tab:traintoken}, increasing the number of reflection tokens from 16 to 128 consistently improves performance across all benchmarks. However, further increasing the number of tokens beyond this point leads to marginal gains or slight degradation. Excessively long reflection sequences might introduce redundancy and make optimization more difficult.

\begin{table*}[h]
\centering
\begin{adjustbox}{width=\linewidth,center}
\renewcommand{\arraystretch}{1.05}
\begin{tabular}{lccccccc}
\toprule  \textbf{\# \texttt{<VR>}} & \textbf{16 tokens} & \textbf{32 tokens} & \textbf{64 tokens} & \textbf{128 tokens} & \textbf{196 tokens} & \textbf{256 tokens} \\
\midrule
V$^{*}$~\cite{wu2024v} & 81.6 & 81.2 & 82.2 & \textbf{84.9} & 83.2 & 83.7\\
HRBench-4K~\cite{wang2025divide} & 71.4 & 72.5 & 73.3 & \textbf{73.8} & 73.5 & 73.7\\
HRBench-8K~\cite{wang2025divide} & 67.8 & 68.4 & 69.9 & 70.1 & \textbf{70.3} & 70.1 \\
\bottomrule
\end{tabular}
\end{adjustbox}
\caption{\textbf{Number of \texttt{<VR>} tokens during training.} We control the maximum number of visual reflection tokens that the model is required to generate.}
\label{tab:traintoken}
\end{table*}

\noindent\textbf{Number of \texttt{<VR>} Tokens during Inference.} We modify the inference procedure by fixing the number of visual reflection tokens (number of \texttt{<VR>}) and explicitly appending the \texttt{<EOR>} token to terminate the reflection phase. The \texttt{<EOR>} token serves as a trigger that forces the model to transition to the final-answer generation stage.

\begin{figure}[h]
    \centering
    \includegraphics[width=\textwidth]{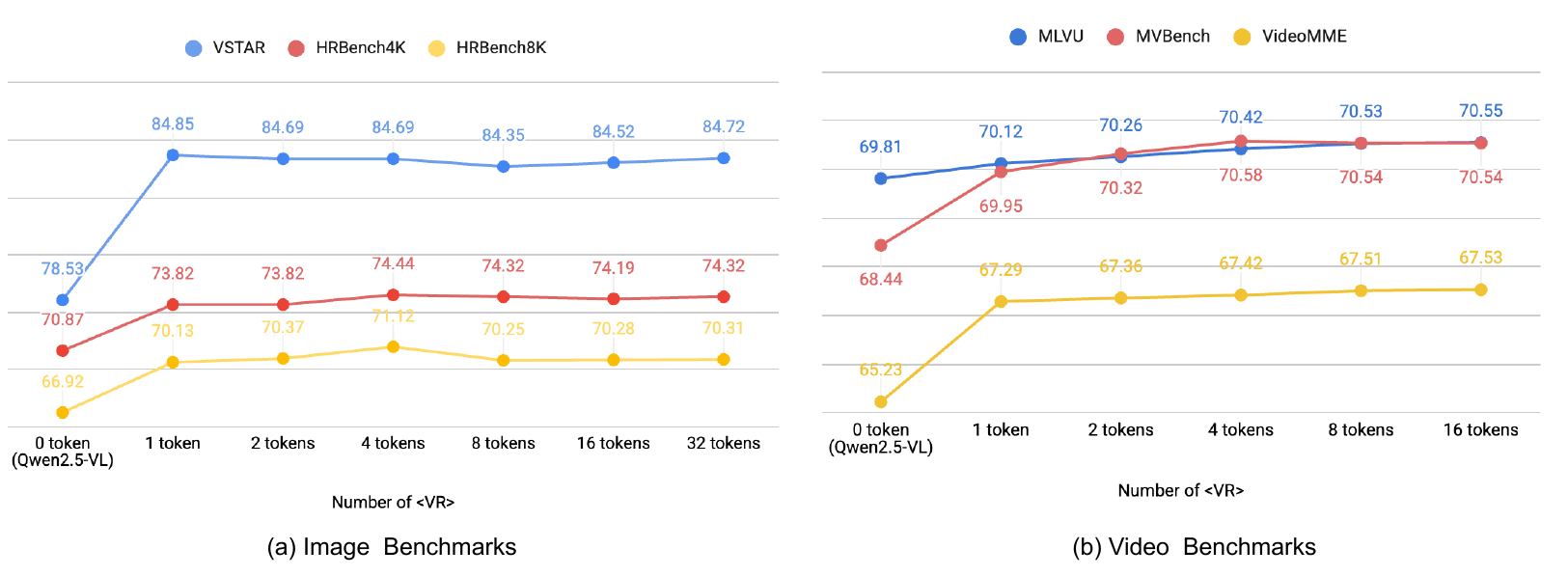}
    \caption{\textbf{Number of \texttt{<VR>} during inference.} We vary the number of generated visual reflection tokens during inference to evaluate how reflection length influences model performance.}
    \label{fig:infertoken}
\end{figure}

As shown in Fig.~\ref{fig:infertoken}, generating one \texttt{<VR>} token already leads to considerable improvements over the backbone Qwen2.5-VL. For image benchmarks, increasing the number of tokens brings limited additional benefit. This suggests that one visual reflection token is sufficient to guide attention toward the relevant small regions in high-resolution images. Therefore, we set the number of \texttt{<VR>} tokens as one for image evaluation.

For video benchmarks, we observe a gradual improvement as the number of reflection tokens increases. However, the gains become marginal when further increasing the token number. Considering the trade-off between performance and computational cost, we use four \texttt{<VR>} tokens for video evaluation.

\paragraph{Attention refocusing, not visual reconstruction.}
Although training employs multiple reflection tokens to encourage comprehensive alignment between reflection embeddings and visual features, we find that only a small number of tokens are necessary at inference time, which effectively guides the model’s attention toward question-relevant visual content. This observation suggests that the visual reflection mechanism primarily acts as an attention refocusing signal, steering the model toward salient regions in images or informative frames in videos, rather than requiring dense reconstruction of the entire set of visual tokens.

\subsection{Analysis}

\noindent\textbf{Inference Efficiency.}
For image VQA, inference is dominated by LLM decoding, making single-pass methods comparable in runtime. We report per-image latency in Table~\ref{tab:imagespeed}. VisReflect remains efficient with a single reflection token while achieving stronger performance.

Table~\ref{tab:zoom} compares our method with zooming-based video models that require multiple prefilling. ``Time'' denotes the average inference time per video. Our method achieves comparable accuracy while significantly reducing inference time, cutting latency by 44\% across benchmarks. This demonstrates that our proposed visual reflection provides a computationally efficient alternative to multi-pass zooming strategies.

\begin{table}[h]
\centering
\begin{adjustbox}{width=\linewidth,center}
\setlength{\tabcolsep}{0.3mm}
\begin{tabular}{lccccccc}
\toprule  \textbf{Model} & PixelReasoner & PAPO & DeepEyes (S) & DeepEyes (M) & Thyme (S) & Thyme (M) & Ours \\
\midrule
Seconds & 3.04 & 2.51 & 2.53 & 4.78 & 2.74 & 4.81 & 2.60 \\
\bottomrule
\end{tabular}
\end{adjustbox}
\caption{\textbf{Inference speed on image understanding.} S denotes single-prefilling (forward pass), while M denotes multiple-prefilling (forward pass).}
\label{tab:imagespeed}
\end{table}

\noindent\textbf{Attention quantification.} We define the ROI attention ratio as $\frac{\frac{1}{|\mathrm{ROI}|}\sum_{i \in \mathrm{ROI}} a_i}{\frac{1}{N}\sum_{i=1}^{N} a_i}$,
where $a_i$ denotes the attention weight of visual token $i$, and $N$ is the total number of visual tokens. A ratio greater than 1 indicates that attention is more concentrated in important regions than across the full image. As show in Table~\ref{tab:attnquan}, our method achieves a higher ratio on V$^{*}$, indicating more effective refocusing toward salient visual regions.

\begin{table}[h]
\centering
\begin{adjustbox}{width=\linewidth,center}
\setlength{\tabcolsep}{0.7mm}
\begin{tabular}{lcccccc}
\toprule
\textbf{Model} & Qwen2.5-VL & PixelReasoner & PAPO & DeepEyes (S) & Thyme (S)  & Ours \\
\midrule
Ratio & 0.73 & 0.75 & 0.81 & 0.79 & 0.82 & 1.15 \\
\bottomrule
\end{tabular}
\end{adjustbox}
\caption{\textbf{Attention quantification.} S denotes single-prefilling (forward pass). A higher ratio indicates that attention is more concentrated in important regions.}
\label{tab:attnquan}
\end{table}

\begin{figure}[h]
    \centering
    \includegraphics[width=\textwidth]{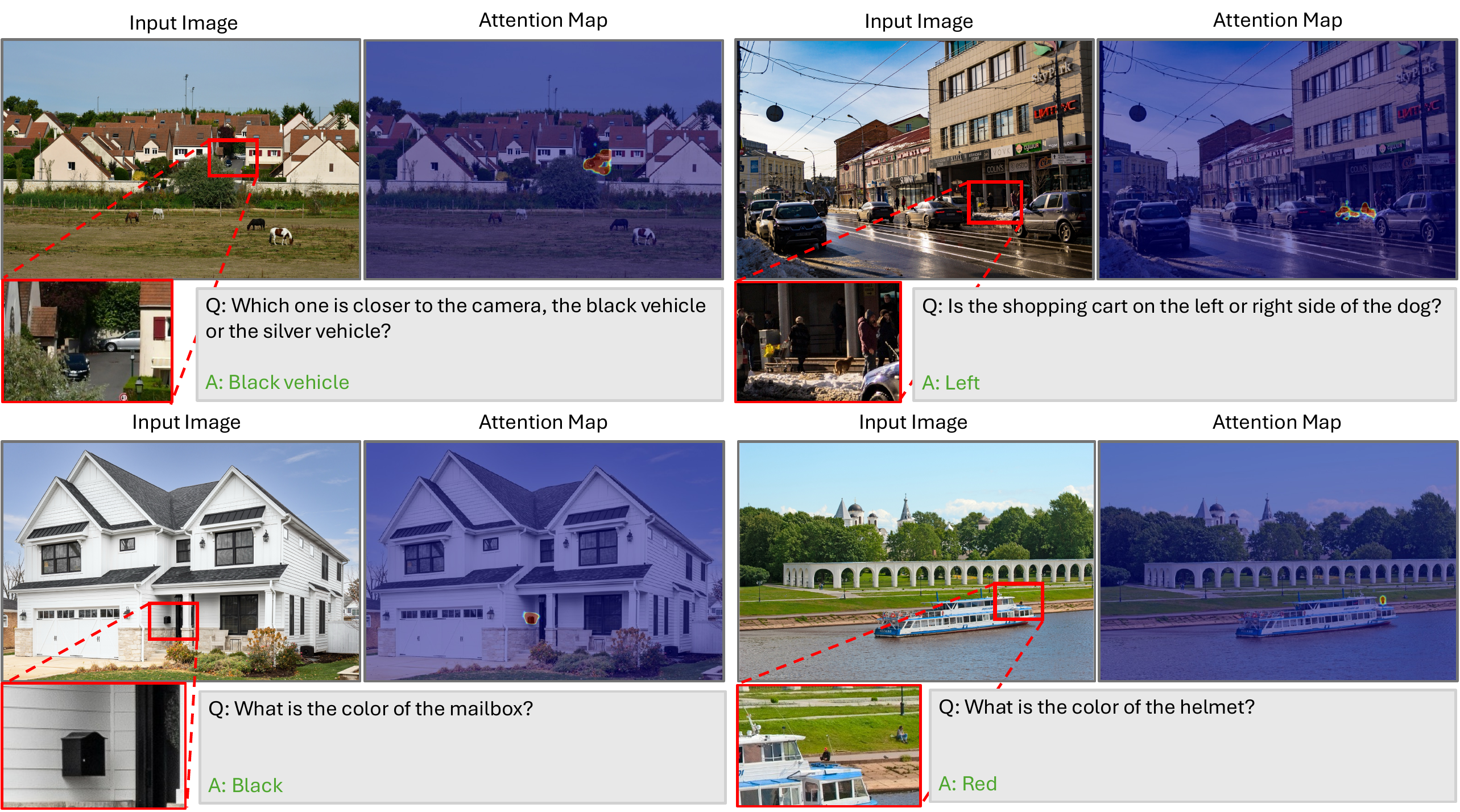}
    \caption{\textbf{Attention visualization of visual reflection tokens (\texttt{<VR>}).} It illustrates the spatial focus of the generated visual reflection tokens on the input image.}
    \label{fig:attn}
\end{figure}
\noindent\textbf{Attention Visualization.}
To better understand how visual reflection guides the model's perception, we visualize the attention maps from the generated \texttt{<VR>} tokens to the input images. As shown in Fig.~\ref{fig:attn}, the attention of reflection tokens consistently concentrates on the question-relevant regions. Notably, these objects often occupy only a small portion of the high-resolution image. Despite the large number of visual tokens, the reflected representations guide the model to selectively focus on informative regions rather than being distracted by irrelevant background content. This observation suggests that visual reflection effectively mitigates the attention sink phenomenon caused by massive visual tokens, enabling the model to better localize and perceive about fine-grained visual details. 

\section{Limitations and Discussion}

During training, the frequency of \texttt{<VR>} tokens is higher than that of the \texttt{<EOR>} token. As a result, the model occasionally tends to continue generating reflection tokens rather than terminating the reflection phase. Therefore, we explicitly insert the \texttt{<EOR>} token after a predefined number of \texttt{<VR>} tokens. The number of reflection tokens should adapt dynamically to the complexity and richness of the question-relevant visual features. Developing an adaptive reflection mechanism is a potential direction for future work.

The video training aligns reflection tokens with visual features at the frame level. However, large-scale video question answering datasets with precise temporal annotations remain limited, and spatiotemporal grounding annotations for QA are even rarer. Richer video datasets with fine-grained spatiotemporal annotations could further enhance the effectiveness of latent visual reflection for long-video perception.

\section{Conclusion}

In conclusion, we presented \textbf{VisReflect}, a simple and principled framework to enhance fine-grained perception in large visual contexts for high-resolution images and long videos. Unlike prior multi-pass paradigms that rely on deterministic token predictions and repeated cropping or zooming, VisReflect generates continuous visual reflection embeddings that selectively emphasize salient regions or frames in latent space. This approach enables efficient single-pass reasoning, guiding cross-modal attention toward relevant visual features without the need for explicit coordinate prediction or iterative re-encoding. Extensive experiments on challenging image benchmarks and long-video benchmarks demonstrate that VisReflect consistently improves performance, achieving an average gain of $4.1\%$ on images and $1.8\%$ on videos compared to strong single-pass baselines. Furthermore, our method reduces inference time by roughly 44\% relative to multi-pass approaches while maintaining comparable accuracy.

\bibliographystyle{splncs04}
\bibliography{main}

\appendix

\title{VisReflect \\-- Supplementary Materials --}

\titlerunning{VisReflect}

\author{ }

\authorrunning{X. Shen et al.}

\institute{ }

\maketitle

The supplementary material provides: 
\begin{itemize}
  \item Section~\ref{sec:lambda}: Effect of weighting factor $\lambda$.
  \item Section~\ref{sec:j}: Number of sampling frames $J$.
  \item Section~\ref{sec:vr}: Number of \texttt{<VR>} tokens.
  \item Section~\ref{sec:coconut}: Comparison with Coconut.
  \item Section~\ref{sec:adaptability}: Inference adaptability.
  \item Section~\ref{sec:attn}: Attention visualization.
  \item Section~\ref{sec:detail}: Implementation details.
\end{itemize}

\section{Effect of weighting factor $\lambda$}
\label{sec:lambda}

The overall training objective consists of two components: the language modeling loss and the visual reflection alignment loss. Specifically, the total loss is defined as $\mathcal{L} = \mathcal{L}_{\mathrm{LM}} + \lambda \mathcal{L}_{\mathrm{VR}}$,
where $\mathcal{L}_{\mathrm{LM}}$ denotes the standard cross-entropy loss for answer generation, and $\mathcal{L}_{\mathrm{VR}}$ enforces alignment between the visual reflection tokens and the corresponding visual representations. The hyperparameter $\lambda$ controls the relative importance of linguistic supervision and latent visual alignment during training. 

Table~\ref{tab:lambda} presents an ablation study on different values of $\lambda$, illustrating how the balance between the two objectives affects overall performance.

\begin{table*}[h]
\centering
\begin{adjustbox}{width=0.9\linewidth,center}
\begin{tabular}{lcccccc}
\toprule  \textbf{$\lambda$} & \, \textbf{V$^{*}$} & \textbf{HRBench-4K} & \textbf{HRBench-8K} & \textbf{MLVU} & \textbf{MVBench} \\
\midrule
0.1 &\, 84.9 & 73.8 & 70.1 & 70.6 & 70.4\\
0.3 &\, 85.2 & 71.6 & 69.1 & 69.2 & 69.3\\
0.5 &\, 84.9 & 71.9 & 69.5 & 69.0 & 68.9\\
0.7 &\, 84.6 & 73.2 & 69.6 & 69.4 & 69.2\\
\bottomrule
\end{tabular}
\end{adjustbox}
\caption{\textbf{Effect of number of factor $\lambda$.}}
\label{tab:lambda}
\end{table*}

\section{Number of sampling frames $J$}
\label{sec:j}

For a video input, the encoder produces frame-wise spatial tokens. Given temporal boundaries $[t_s, t_e]$, we uniformly sample $J$ frames with indices $t_1, t_2, \dots, t_J$, where each $t_j \in [t_s, t_e]$. In Table~\ref{tab:j} we investigate how many video frames are needed for visual reflection.

\begin{table*}[h]
\centering
\begin{adjustbox}{width=0.95\linewidth,center}
\begin{tabular}{lcccc}
\toprule  \textbf{$J$} & \textbf{MLVU} & \textbf{MVBench} & \textbf{VideoMME (w/o sub.)} & \textbf{VideoMME (w/ sub.))} \\
\midrule
1 & 70.0 & 69.4 & 66.8 & 71.6\\
2 & 70.3 & 69.6 & 67.2 & 71.9\\
3 & 70.6 & 70.4 & 67.4 & 72.7\\
4 & 70.2 & 70.2 & 67.6 & 72.6\\
5 & 70.4 & 69.9 & 67.1 & 72.0\\
\bottomrule
\end{tabular}
\end{adjustbox}
\caption{\textbf{Effect of number of sampling frames $J$.}}
\label{tab:j}
\end{table*}

\section{Number of \texttt{<VR>} tokens}
\label{sec:vr}

We modify the inference procedure by fixing the number of visual reflection tokens (number of \texttt{<VR>}) and explicitly appending the \texttt{<EOR>} token to terminate the reflection phase. The \texttt{<EOR>} token serves as a trigger that forces the model to transition to the final-answer generation stage.

\noindent\textbf{Number of \texttt{<VR>} Tokens on Image Understanding.} As shown in Table~\ref{tab:infertokenimage}, generating one \texttt{<VR>} token already leads to considerable improvements over the backbone Qwen2.5-VL. For image benchmarks, increasing the number of tokens brings limited additional benefit. This suggests that one visual reflection token is sufficient to guide attention toward the relevant small regions in high-resolution images. Therefore, we set the number of \texttt{<VR>} tokens as one for image evaluation.

\begin{table*}[h]
\centering
\begin{adjustbox}{width=\linewidth,center}
\begin{tabular}{lccccccccc}
\toprule  \textbf{\# \texttt{<VR>}} & \textbf{0 token (Qwen2.5-VL)} & \textbf{1 tokens} & \textbf{2 tokens} & \textbf{4 tokens} & \textbf{8 tokens} & \textbf{16 tokens} & \textbf{32 tokens} \\
\midrule
V$^{*}$~\cite{wu2024v} & 78.53 & 84.85 & 84.69 & 84.69 & 84.35 & 84.52 & 84.72\\
HRBench-4K~\cite{wang2025divide} & 70.87 & 73.82 & 73.82 & 74.44 & 74.32 & 74.19 & 74.32\\
HRBench-8K~\cite{wang2025divide} & 66.92 & 70.13 & 70.37 & 71.12 & 70.25 & 70.28 & 70.31 \\
\bottomrule
\end{tabular}
\end{adjustbox}
\caption{\textbf{Effect of the number of \texttt{<VR>} tokens during inference on image understanding benchmarks.}}
\label{tab:infertokenimage}
\end{table*}

\noindent\textbf{Number of \texttt{<VR>} Tokens on Video Understanding.} At Table~\ref{tab:infertokenvideo}, we observe a gradual improvement as the number of reflection tokens increases. However, the gains become marginal when further increasing the token number. Considering the trade-off between performance and computational cost, we use four \texttt{<VR>} tokens for video evaluation.

\begin{table*}[h]
\centering
\begin{adjustbox}{width=\linewidth,center}
\begin{tabular}{lccccccccc}
\toprule  \textbf{\# \texttt{<VR>}} & \textbf{0 token (Qwen2.5-VL)} & \textbf{1 tokens} & \textbf{2 tokens} & \textbf{4 tokens} & \textbf{8 tokens} & \textbf{16 tokens} \\
\midrule
MLVU~\cite{zhou2025mlvu} & 69.81 & 70.12 & 70.26 & 70.42 & 70.53 & 70.55 \\
MVBench~\cite{li2024mvbench} & 68.44 & 69.95 & 70.32 & 70.58 & 70.54 & 70.54 \\
VideoMME (w/o sub.)~\cite{fu2025video} & 65.23 & 67.29 & 67.36 & 67.42 & 67.51 & 67.53 \\
\bottomrule
\end{tabular}
\end{adjustbox}
\caption{\textbf{Effect of the number of \texttt{<VR>} tokens during inference on video understanding benchmarks.}}
\label{tab:infertokenvideo}
\end{table*}

\section{Comparison with Coconut}
\label{sec:coconut}

Coconut~\cite{coconut} (i) operates purely in the language space and (ii) without supervision over latent states. We adapt it to the visual setting for comparison. It underperforms our method as shown in Table~\ref{tab:coconut}.

\begin{table}[h]
\centering
\begin{adjustbox}{width=\linewidth,center}
\renewcommand{\arraystretch}{0.85}
\setlength{\tabcolsep}{2.7mm}
\begin{tabular}{lcccc}
\toprule  
\textbf{Model} & BLINK & HRBench-4K & HRBench-8K & V* \\
\midrule
Coconut & 56.2 & 71.5 & 67.4 & 79.4 \\
VisReflect (Ours) & 57.1 & 73.8 & 70.1 & 84.9\\
\bottomrule
\end{tabular}
\end{adjustbox}
\caption{Comparison with Coconut.}
\label{tab:coconut}
\end{table}

\section{Inference Adaptability.}
\label{sec:adaptability}

We apply a threshold (14.5) to the logits of \texttt{<VR>} to dynamically determine whether additional reflection tokens should be generated. As shown below in Table~\ref{tab:adaptability}, adaptive reflection lengths better accommodate varying question complexity.

\begin{table}[h]
\centering
\begin{adjustbox}{width=\linewidth,center}
\begin{tabular}{lcccccccc}
\toprule
\multirow{2}{*}{\textbf{Methods}}
& \multicolumn{2}{c}{\textbf{BLINK}}
& \multicolumn{2}{c}{\textbf{HRBench-4K}}
& \multicolumn{2}{c}{\textbf{HRBench-8K}}
& \multicolumn{2}{c}{\textbf{V$^{*}$}} \\
\cmidrule(lr){2-3} \cmidrule(lr){4-5} \cmidrule(lr){6-7} \cmidrule(lr){8-9} & Acc & \# Tokens & Acc & \# Tokens & Acc & \# Tokens & Acc & \# Tokens \\
\midrule
Fix & 57.1 & 1.0 & 73.8 & 1.0 & 70.1 & 1.0 & 84.9 & 1.0 \\
Adaptive & 57.2 & 2.3 & 74.2 & 3.7 & 71.2 & 4.3 & 85.8 & 2.6\\
\bottomrule
\end{tabular}
\end{adjustbox}
\caption{Inference Adaptability.}
\label{tab:adaptability}
\end{table}

\section{Attention Visualization}
\label{sec:attn}

\begin{figure}[htbp]
    \centering
    \includegraphics[width=0.8\textwidth]{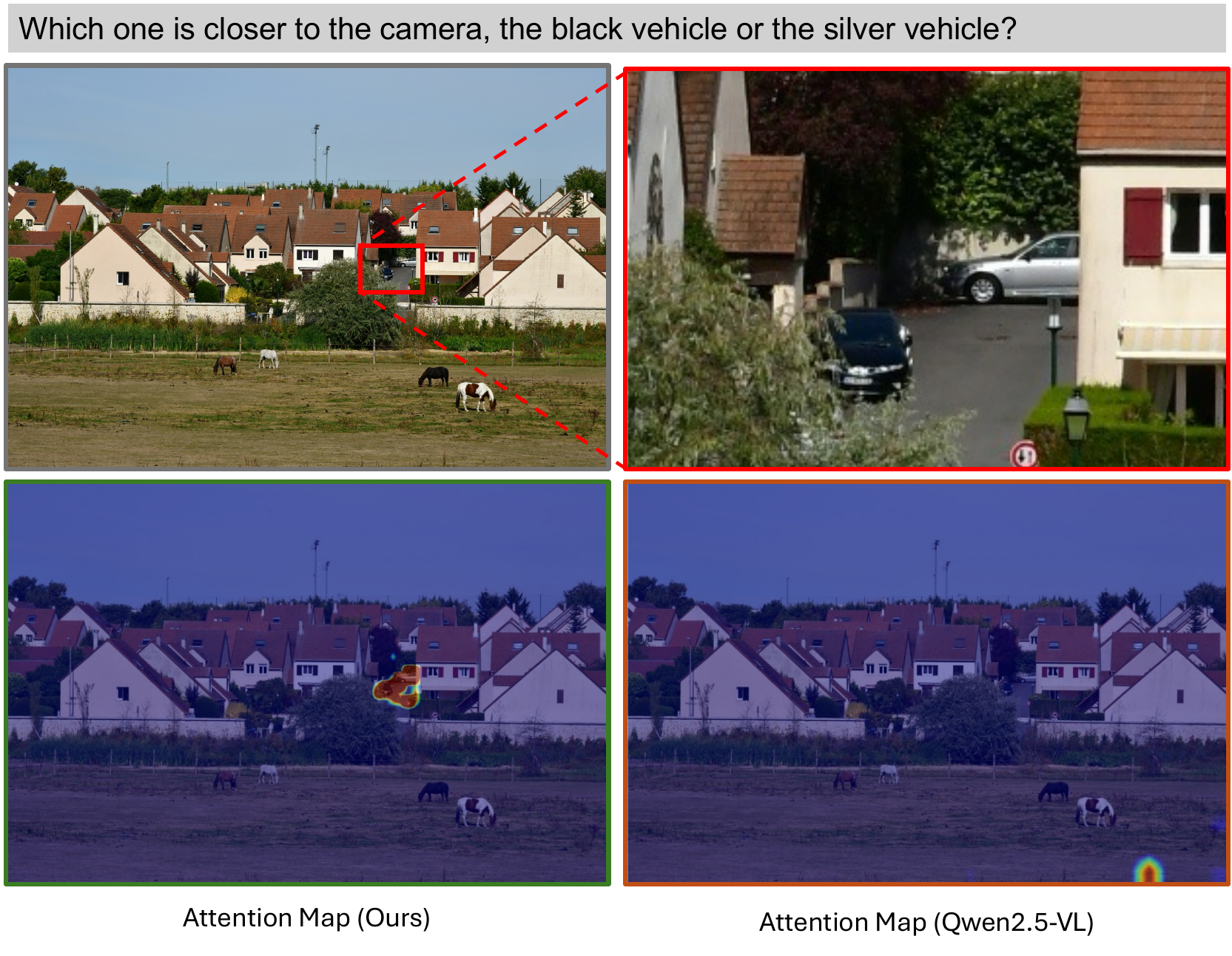}
    \caption{Attention map visualization comparing our VisReflect with the baseline Qwen2.5-VL.}
    \label{fig:qual1}
\end{figure}

\begin{figure}[htbp]
    \centering
    \includegraphics[width=0.8\textwidth]{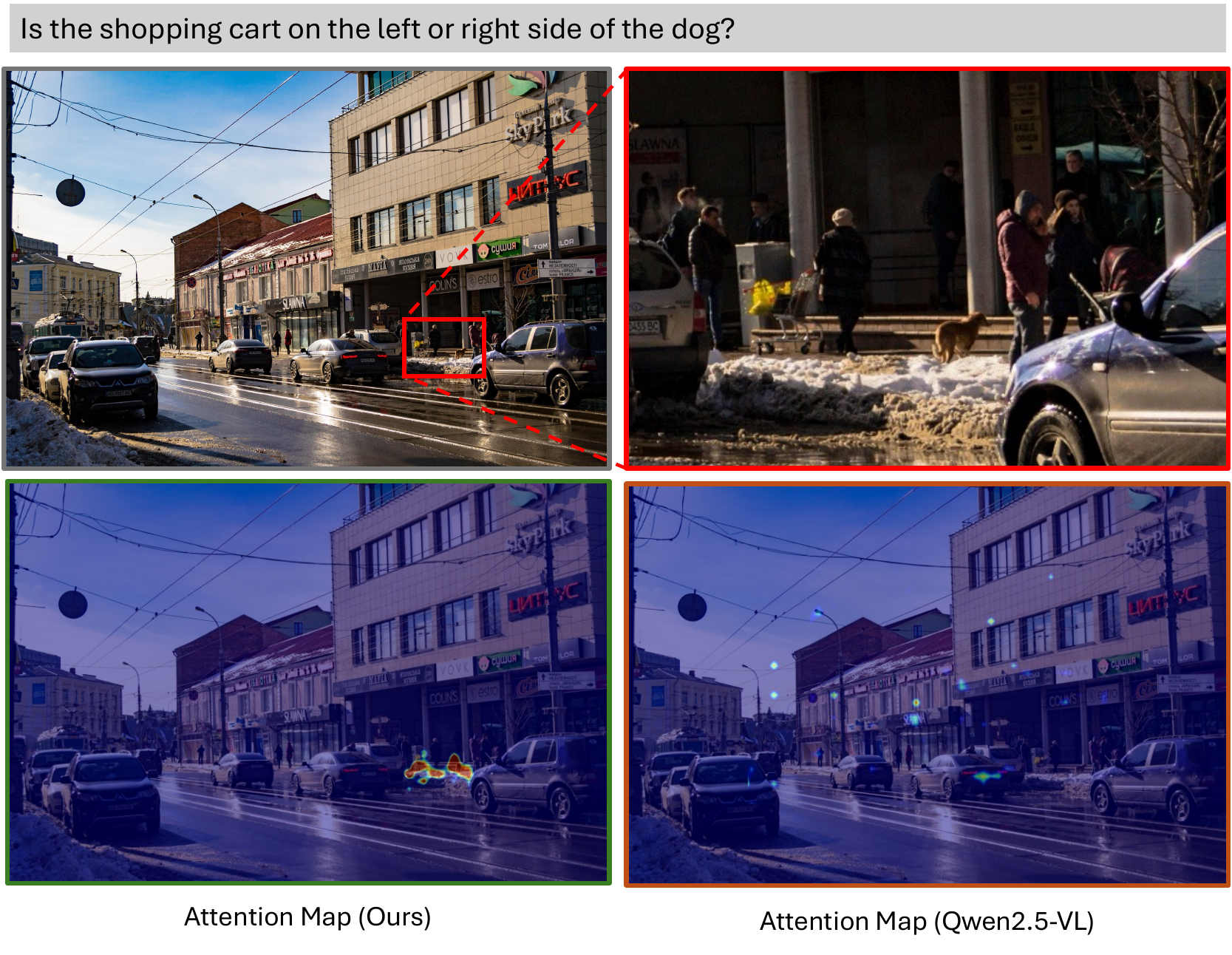}
    \caption{Attention map visualization comparing our VisReflect with the baseline Qwen2.5-VL.}
    \label{fig:qual2}
\end{figure}

\begin{figure}[htbp]
    \centering
    \includegraphics[width=0.8\textwidth]{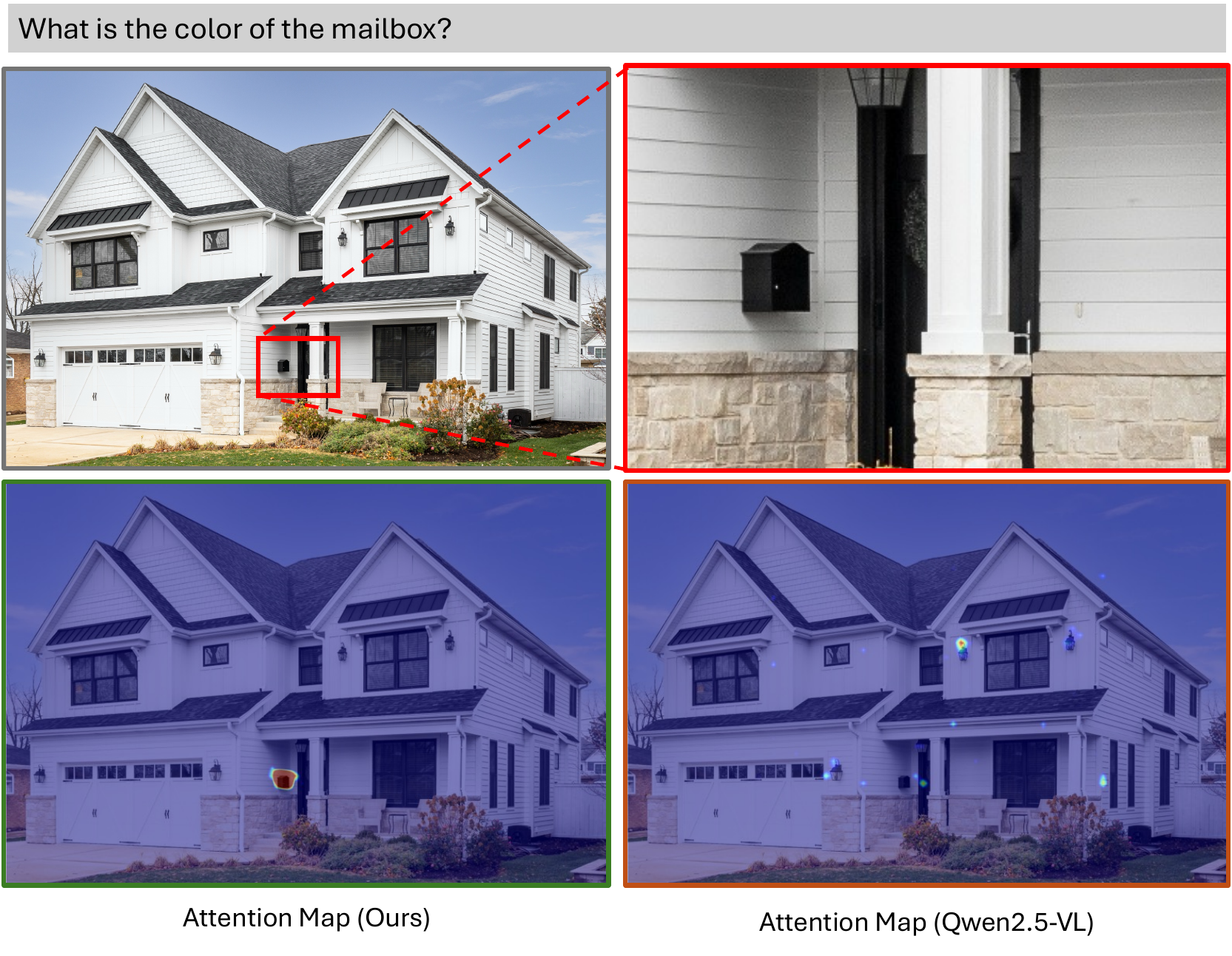}
    \caption{Attention map visualization comparing our VisReflect with the baseline Qwen2.5-VL.}
    \label{fig:qual3}
\end{figure}

\begin{figure}[htbp]
    \centering
    \includegraphics[width=0.8\textwidth]{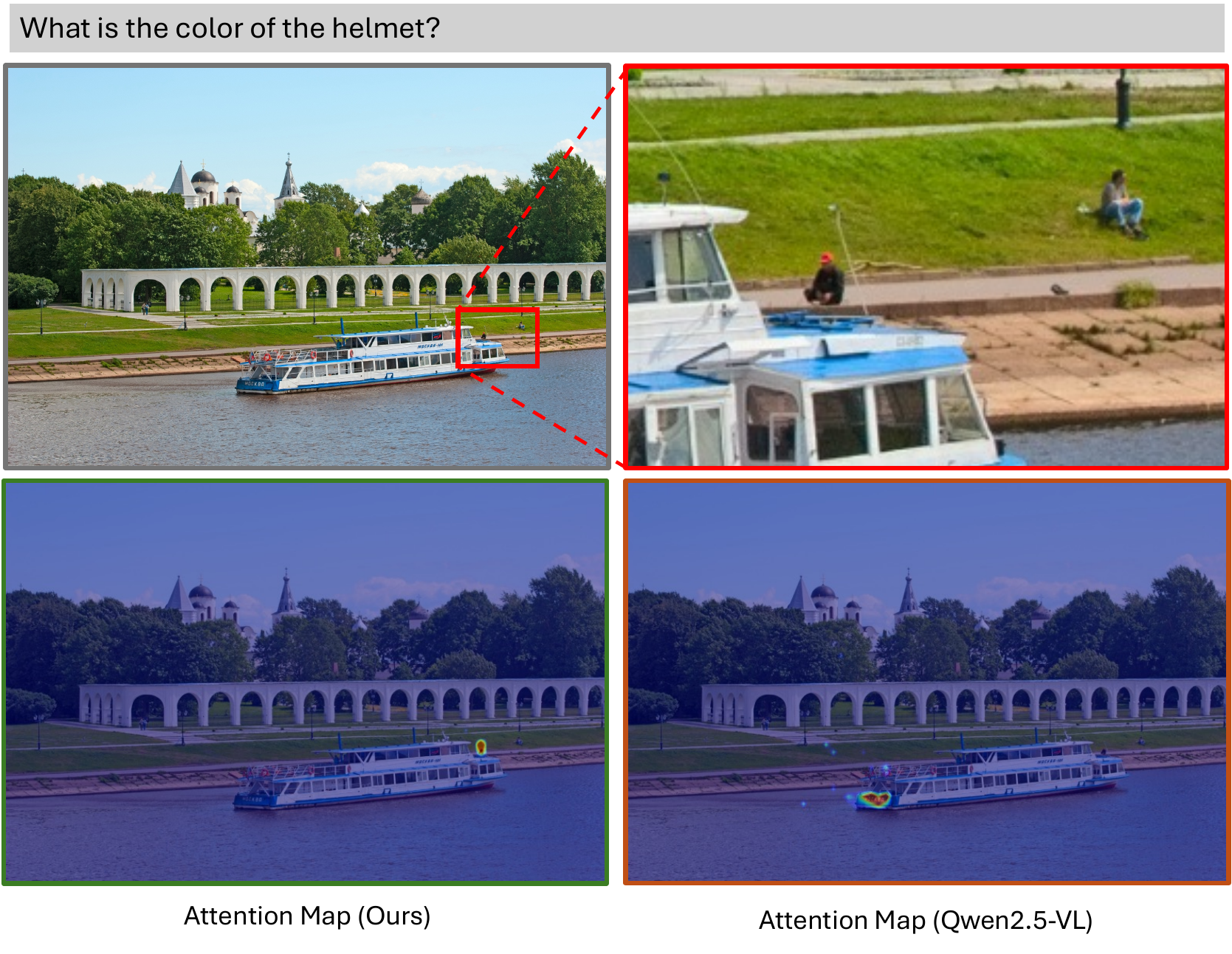}
    \caption{Attention map visualization comparing our VisReflect with the baseline Qwen2.5-VL.}
    \label{fig:qual4}
\end{figure}

To better understand how visual reflection guides the model’s perception, we visualize the attention maps from the generated \texttt{<VR>} tokens to the input images and compare them with those of the base model, Qwen2.5-VL, which does not include visual reflection. As shown in Fig.~\ref{fig:qual1}--\ref{fig:qual4}, the attention of the reflection tokens consistently concentrates on question-relevant regions. In contrast, the base model tends to distribute attention over large irrelevant areas, failing to focus on the target objects. This observation suggests that visual reflection effectively mitigates the attention sink phenomenon caused by massive visual tokens, enabling the model to better localize and perceive fine-grained visual details.

\section{Implementation Details}
\label{sec:detail}

We adopt Qwen2.5-VL-7B~\cite{bai2025qwen25vl} as the backbone model. The training procedure is organized into two stages, starting from image-based supervision and subsequently extending to video data. This staged training strategy stabilizes optimization and enables the model to first learn localized visual grounding from images before adapting to temporal reasoning in videos.

\noindent\textbf{Stage 1: Image Training.}
In the first stage, we train the model using image datasets to learn spatial visual reflection. The maximum input resolution is set to $4096 \times 28 \times 28$ pixels, while the minimum resolution is $128 \times 28 \times 28$ pixels. Under the visual tokenizer used by the backbone model, this corresponds to a visual token range of $128 \leq N \leq 4096$. Varying the resolution during training exposes the model to different levels of visual granularity and improves robustness to diverse visual contexts.

For supervision, we employ the VisualCoT dataset~\cite{shao2024visual}. Since our objective is to encourage the model to focus on localized visual evidence, we filter the training samples to retain only those whose annotated bounding box area occupies less than 10\% of the entire image. This filtering step biases the training distribution toward cases requiring fine-grained spatial perception and discourages trivial global reasoning.

\noindent\textbf{Stage 2: Video Training.}
In the second stage, we initialize the model from the checkpoint obtained in Stage 1 and continue training on video datasets to learn temporal visual reflection. The maximum number of visual tokens is increased to $16{,}384$ to accommodate long video sequences. Meanwhile, to control memory usage, the maximum pixel resolution per frame is limited to $128 \times 28 \times 28$.

For each training example, we sample $J=3$ visual reflection frames from the annotated temporal span that contains the relevant visual evidence. These frames serve as supervision signals to guide the model in identifying temporally salient visual content.

The training datasets include QVHighlights~\cite{lei2021detecting}, ActivityNet~\cite{yu2019activitynet}, NExT-GQA~\cite{xiao2024can}, and the multiple-choice split of PLM-Video~\cite{cho2025perceptionlm}. To emphasize localized temporal reasoning, we further constrain the annotated temporal clue span to be shorter than $10$ seconds.

During both training stages, the visual encoder and multimodal projector remain frozen, and only the parameters of the language model are updated. This design preserves the pretrained visual representations while enabling the language model to learn the latent visual reflection mechanism.

We train the model using the AdamW optimizer with a learning rate of $1\times10^{-5}$ and a warmup ratio of $0.03$. The alignment loss is weighted by $\lambda = 0.1$ to balance it with the language modeling objective. All experiments are conducted using 8 NVIDIA H100 GPUs (80GB).

\end{document}